\documentclass{article}

\usepackage{microtype}
\usepackage{graphicx}
\usepackage{subfigure}
\usepackage{booktabs} 
\usepackage{tikz}
\usetikzlibrary{shapes, arrows.meta, positioning}
\usepackage{hyperref}
\usepackage[table]{xcolor}

\usepackage[utf8]{inputenc}
\DeclareUnicodeCharacter{0002}{}

\usepackage[accepted]{icml2026}
\setcitestyle{numbers,square,sort&compress,comma}

\usepackage{amsmath}
\usepackage{amssymb}
\usepackage{mathtools}
\usepackage{amsthm}

\usepackage{multirow}
\usepackage[most]{tcolorbox}

\usepackage{caption}
\usepackage{enumitem}

\definecolor{promptframe}{RGB}{154,164,201}
\definecolor{promptback}{RGB}{248,249,252} 

\definecolor{boxbg}{RGB}{245,247,251}        
\definecolor{boxframe}{RGB}{165,174,208}     
\definecolor{headerbg}{RGB}{232,236,246}     
\definecolor{divider}{RGB}{176,184,214}      
\definecolor{titletext}{RGB}{95,105,150}

\tcbset{
  comparisonbox/.style={
    colback=boxbg,
    colframe=boxframe,
    boxrule=0.6pt,
    arc=3pt,
    left=8pt,
    right=8pt,
    top=6pt,
    bottom=8pt,
    enhanced,
    colbacktitle=headerbg,
    coltitle=black,
    fonttitle=\bfseries\small,
    attach boxed title to top left={
      yshift=-1mm,
      xshift=6pt
    },
    boxed title style={
      boxrule=0pt,
      arc=2pt,
      left=6pt,
      right=6pt,
      top=3pt,
      bottom=3pt
    }
  }
}

\usepackage[capitalize,noabbrev]{cleveref}

\theoremstyle{plain}

\theoremstyle{definition}

\theoremstyle{remark}

\usepackage[textsize=tiny]{todonotes}

\begin{document}

\twocolumn[
\icmltitle{Is Video Anomaly Detection Misframed? \\ Evidence from LLM-Based and Multi-Scene Models}

\vskip 0.15in

\begin{center}
{\fontsize{11.25pt}{13.5pt}\bfseries\selectfont
Furkan Mumcu$^{\dagger, 1}$ \quad
Michael J.~Jones$^{*,2}$ \quad
Anoop Cherian$^{*,3}$ \quad
Yasin Yilmaz$^{\dagger,4}$
}

\vskip 0.1in

{\small
$^{\dagger}$University of South Florida \quad
$^{*}$Mitsubishi Electric Research Laboratories (MERL)
}

\vskip 0.08in

{\tt\small
$^{1}$furkan@usf.edu \quad
$^{2}$mjones@merl.com \quad
$^{3}$cherian@merl.com \quad
$^{4}$yasiny@usf.edu
}
\end{center}

\vskip 0.25in
]

\begin{abstract}
Recent video anomaly detection research has expanded rapidly with an emphasis on general models of normality intended to work across many different scenes. While this focus has led to improvements in scalability and multi-scene generalization, it has also shifted the field away from modeling the scene-specific and context-dependent nature of normal behavior. Contemporary approaches frequently rely on video-level weak supervision and opaque pretrained representations from multi-modal large language models (MLLMs), which encourage models to respond to familiar semantic anomaly categories rather than to deviations from the normal patterns of a particular environment. This trend suppresses spatial localization, introduces semantic bias, and reduces anomaly detection to a form of action recognition. In this paper, we examine whether these prevailing formulations align with the core requirements of real-world VAD, which is typically performed within a single scene where normality is determined by local geometry, semantics, and activity patterns. Through targeted visual analyses and empirical evaluations, we demonstrate the practical consequences of these limitations and show that meaningful progress in VAD requires renewed focus on single-scene, spatially-aware, and explainable formulations that capture the nuanced structure of normality within individual environments.
\end{abstract}

\section{Introduction}

\begin{figure}[t]
    \centering
    \begin{minipage}[t]{0.48\columnwidth}
        \centering
        \includegraphics[width=\linewidth]{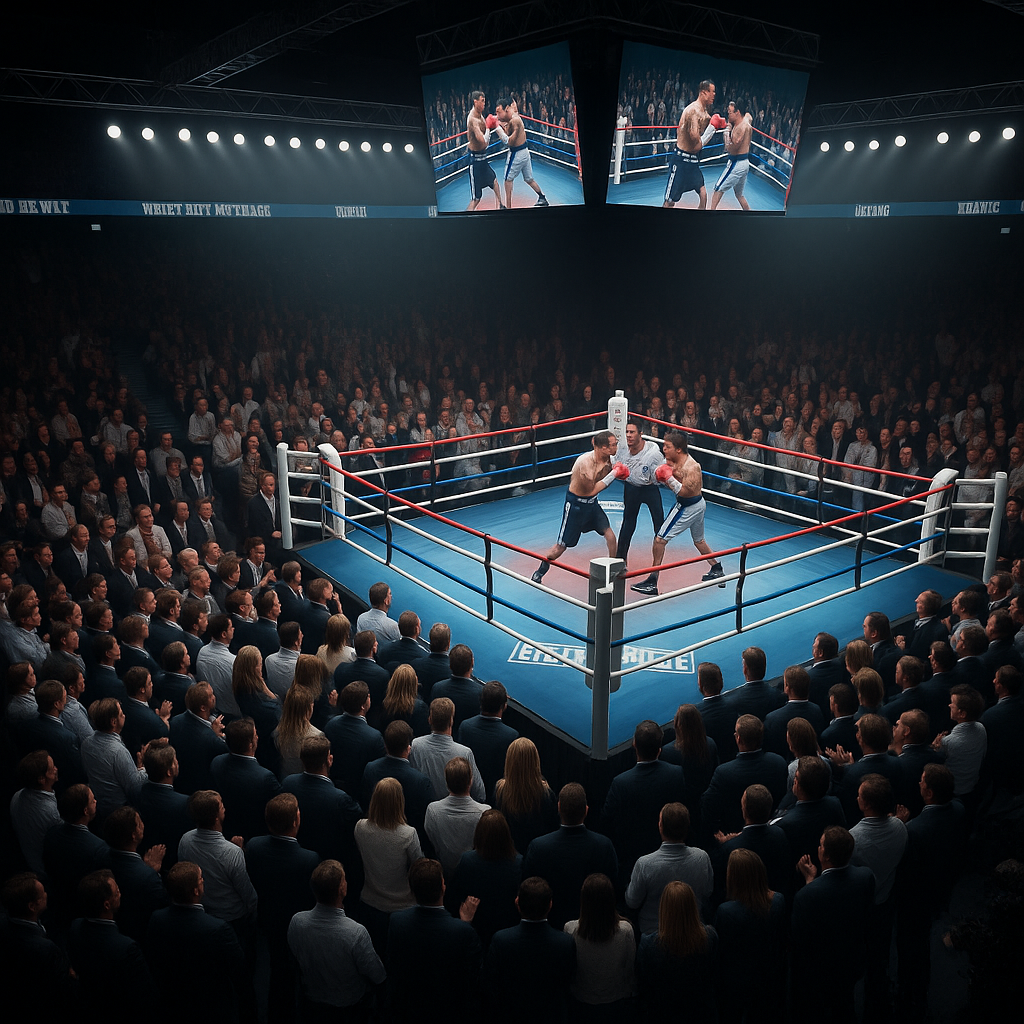}
    \end{minipage}
    \hfill
    \begin{minipage}[t]{0.48\columnwidth}
        \centering
        \includegraphics[width=\linewidth]{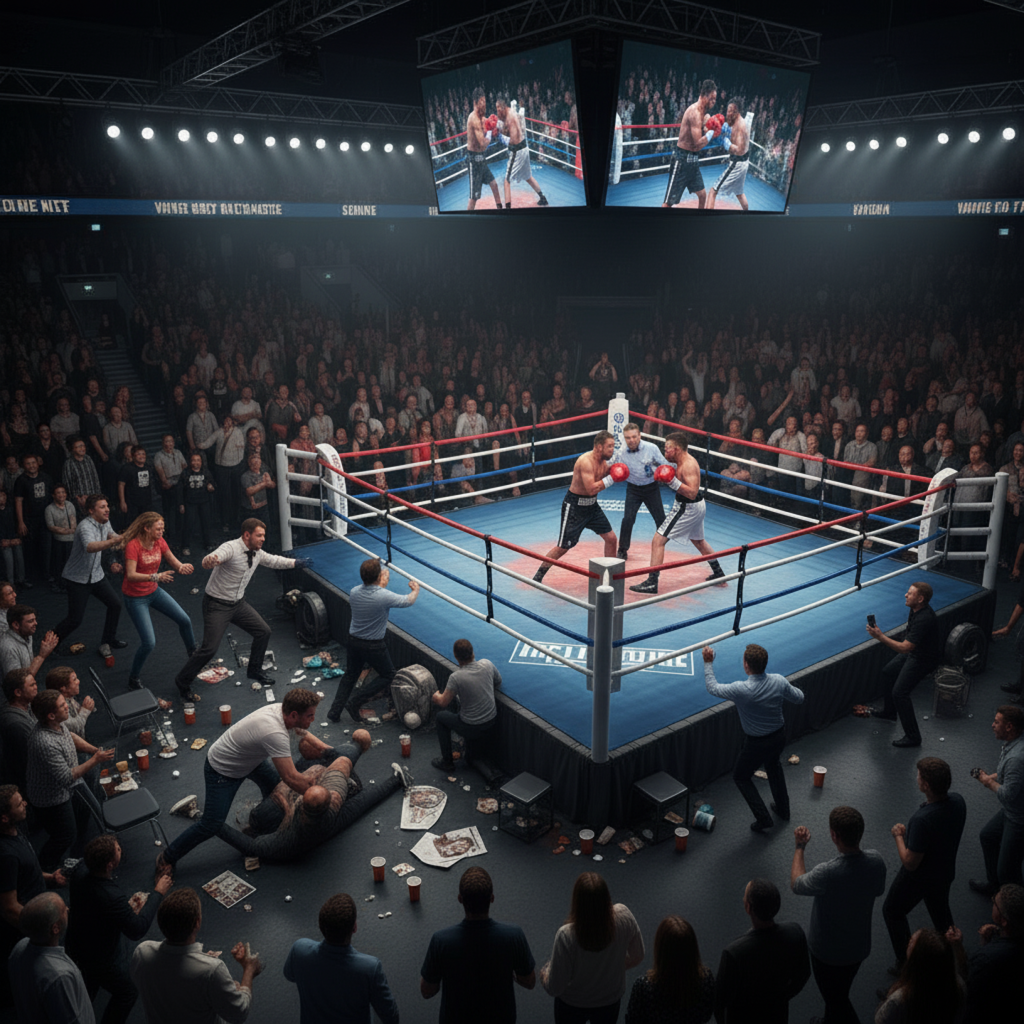}
    \end{minipage}
    \caption{These images illustrate a core limitation of current video anomaly detection approaches. In the first image, the fighting occurs inside the boxing ring, a context where such an action is normal. However, recent models trained primarily with weak supervision or relying on an LLM's built-in notion of normality tend to flag this as anomalous because they focus on the high-level action category rather than the scene context. In contrast, the second image shows another fight among the audience, which constitutes a genuine anomaly in this environment. Yet, current models often fail to differentiate such cases, as they are unable to localize and interpret anomalies according to their spatial and contextual relevance within the scene.}
    \label{fig:intro}
\end{figure}

Video Anomaly Detection (VAD) has become a vital research frontier in intelligent surveillance, safety monitoring, and situational awareness. Its goal is to automatically identify activity in videos that deviate from expected visual patterns, such as a vehicle moving against traffic or a person entering a restricted area. Early research predominantly focused on single-scene formulations, where models learn normal behavioral patterns for a specific camera view \cite{crowded2010,xu2015learning,hasan2016learning}. This setting closely matches real-world deployments, in which each surveillance camera operates within a fixed environment that defines its own spatial and semantic context of normality. Over time, an alternative line of research emerged that explored multi-scene formulations, often employing weakly supervised \cite{sultani2018real} or training-free paradigms \cite{harnessing2024}. In these approaches, models are trained across diverse environments (sometimes using video-level anomaly labels) to learn representations that generalize beyond a single scene. This paradigm shifted the focus of VAD by emphasizing scalability and multi-scene generalization, while reducing emphasis on scene-specific structure and contextual detail.

In recent years, the research landscape of VAD has shifted markedly with the emergence of Large Language Models (LLMs) and Multimodal Large Language Models (MLLMs). These models have introduced new opportunities to integrate semantic understanding and visual reasoning, prompting a rapid expansion of LLM-driven approaches to anomaly detection. This transformation is evident in recent publication patterns across major computer vision and machine learning venues, where most new VAD studies employ weak supervision and multi-scene training schemes or training-free approaches, both built upon pretrained vision–language backbones. 
By 2025, a clear majority of VAD papers in leading conferences adopt multi-scene approaches using weakly-supervised or training-free methods (please see Figure \ref{fig:survey}).

This widespread adoption of approaches that learn general models of normal activity which are applicable across multiple scenes reflects a fundamental redefinition of what constitutes an anomaly in video understanding. Instead of constructing detailed models of scene-specific regularities, recent methods rely on general models intended to detect common anomaly types across diverse scenes. Although this trend facilitates dataset creation and improves scalability, it reduces the modeling of spatial and contextual dependencies that make anomaly detection a uniquely challenging problem. Multi-scene VAD formulations imply that models of normality should generalize across different scenes. While this property may seem beneficial at first glance, it does not always account for the fact that normal activity is often specific to a particular scene. Learning what is normal in one scene may not transfer effectively to another. For example, cars parked in some locations along a street are normal while in other locations they are anomalous. Fighting may be normal in some scenes and anomalous in others (Figure \ref{fig:intro}). Determining what activities are normal in a particular scene generally requires observing that scene directly. Scene-specific and location-specific anomalies are commonplace in many surveillance applications, but are not explicitly modeled by multi-scene VAD algorithms.

Many recent methods that build general models of normality follow the weakly-supervised training paradigm, which further reduces the focus on practical usage scenarios. Weak supervision supplies ground-truth anomaly classes for some training videos without providing ground truth temporal or spatial locations. Furthermore, weakly supervised approaches rarely evaluate generalization beyond the anomaly classes supplied during training. In practice, these properties have two undesirable effects. First, they encourage VAD algorithms to specialize in detecting only the classes of anomalies that occur in ground truth annotations. They are not required to generalize to, nor evaluated on, unseen anomaly types, which are commonly expected in real-world VAD applications. This closed-world formulation shifts weakly-supervised VAD toward a form of action classification, rather than adhering to the open-world nature of practical VAD. Second, the lack of spatial localization in the weak annotations discourages the model from producing spatially localized anomaly detections. Spatial localization is important for explaining anomalies and for drawing a human operator's attention to the anomaly in a cluttered scene.

These observations highlight several key limitations of current VAD paradigms. In particular, general models often struggle to capture scene-specific normal activity and its dependence on spatial and positional context. Approaches based on weak supervision and large pretrained models tend to respond to familiar semantic categories rather than deviations from normal behavior, effectively shifting the problem toward action recognition. In addition, the lack of spatial grounding in many pipelines limits interpretability and practical usability, while reliance on opaque pretrained data raises concerns about evaluation transparency and generalization.

This shift introduces a mismatch between current VAD formulations and real-world deployment settings. In practice, anomaly detection is typically performed within a single camera view, where the goal is to identify deviations from the normal behavioral patterns characteristic of that specific environment. By prioritizing scalability and multi-scene generalization, recent research trends move away from this setting, effectively treating VAD as a video classification problem across heterogeneous scenes. The limitations outlined above (loss of scene-specific modeling, loss of contextual reasoning, semantic bias toward familiar activities, lack of spatial localization, and dependence on opaque pretrained data) are direct consequences of this shift. We therefore emphasize the importance of formulations that focus on single-scene, spatially aware, and explainable modeling of normality within individual environments. Such a focus is essential for advancing toward systems that are practical, robust, interpretable, and capable of being applied to real-world scenarios.

In this paper, we make the following contributions:
\begin{itemize}
\item We identify a fundamental mismatch between prevailing multi-scene VAD formulations and real-world deployment settings, where anomaly detection is inherently scene-specific and context-dependent.
\item We provide empirical evidence demonstrating that recent weakly-supervised and training-free methods degrade significantly when evaluated under single-scene settings.
\item We analyze the role of semantic priors from large pretrained models and show how they bias VAD systems toward recognizing predefined activity categories rather than detecting deviations from normal behavior.
\item We highlight the importance of spatial grounding, interpretability, and transparency, and advocate for formulations that emphasize single-scene, spatially-aware, and explainable modeling of normality.
\end{itemize}

\section{Motivation and Background}

\begin{figure*}[t]
   \centering
    \includegraphics[width=1\linewidth]{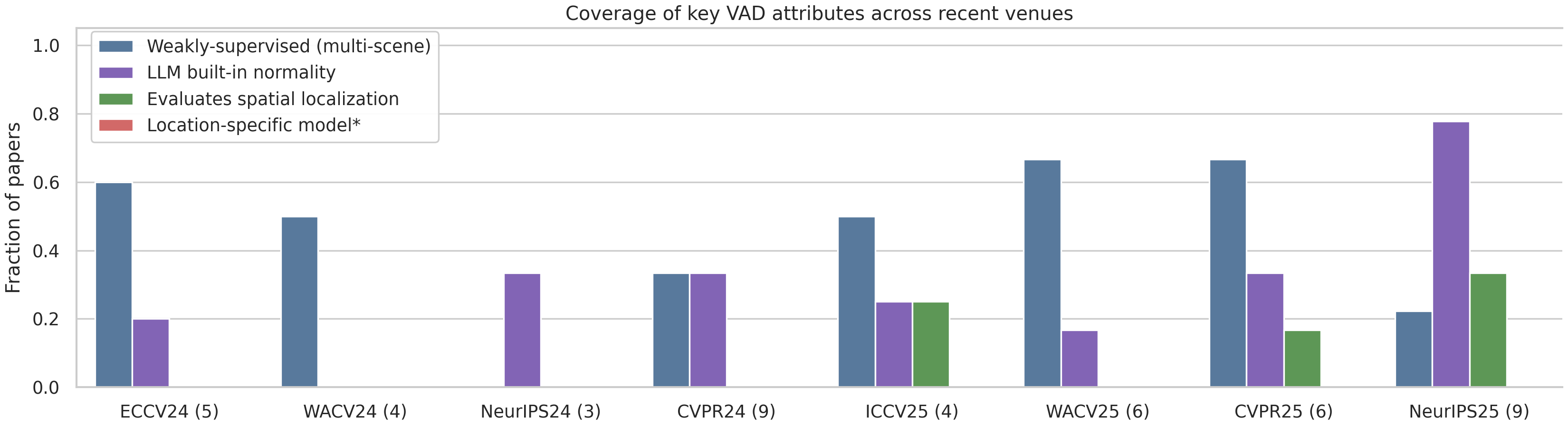}
\caption{Coverage of key video anomaly detection (VAD) attributes across recent venues. 
Numbers in parentheses next to each venue indicate the total number of VAD papers at that venue. 
Location-specific model* denotes methods that explicitly model spatially conditioned normality; no surveyed papers satisfy this criterion, resulting in zero observed coverage.}
    \label{fig:survey}
\end{figure*}

As previously discussed, the goal in single-scene VAD is to build a model of normal activity for a particular scene given normal video of that scene and then to detect activity that significantly differs from normal activity in testing video of the same scene. ``Single-scene" does not necessarily imply ``static-camera", although all previous research that we are aware of that has taken a single-scene approach has also assumed a static camera.  However, a pan-tilt-zoom (PTZ) camera, for example, can also be considered as providing single-scene video even though it only views a part of the scene at a time.  A multi-camera system looking at different parts of a scene could also fit into the single-scene formulation.

In addition to being scene-dependent, the normal model should also be location-dependent.  This is because the same activity may be normal in some locations of the scene, but anomalous in other locations.  For example, people may walk in some areas but not others, cars may drive in some areas but not others, and as Figure \ref{fig:intro} shows, people may fight in some areas but not others.  In multi-scene VAD, because the normal model should generalize across multiple scenes and there are no common locations across scenes, models do not need to be location-dependent.  Location-dependent anomalies are not found in multi-scene datasets.

A model may have context dependence, as well as location dependence.  These are distinct but related concepts.  A context-dependent model may learn that people walk on sidewalks but not on top of cars, or that cars drive on roads but not on sidewalks.  Using a context-dependent model, however, is not sufficient to cover all types of anomalies that can occur in single-scene VAD.  For example, in a particular scene, people might walk on the grass in one region of the scene, but never walk on the grass in another region.  The context of ``grass" is not enough to distinguish normal activity from anomalous.  In that particular scene, it is only the location of the ``walking on grass" activity that distinguishes normal from anomalous.  Similarly, cars may park in some spots along a street, but not in others.  Again, the location is critical for distinguishing normal from anomalous activity.
For these reasons, we argue that location dependence as well as context dependence are necessary attributes for single-scene VAD models to have.

Unfortunately, the older single-scene VAD datasets such as Ped1~\cite{crowded2010}, Ped2~\cite{crowded2010} and Avenue~\cite{matlab2013} contain very few location-dependent anomalies which has had the effect of downplaying the importance of location-dependent modeling since it is not strictly necessary for high accuracy on those datasets.

Another unfortunate trend in VAD research is the lack of spatial evaluation of anomaly detections.  Most papers focus solely on the detection of frames containing anomalies without regard to the correct spatial localization of anomalies.  Spatial localization is important in many applications for a couple of reasons.  First, VAD systems are typically used to alert a human operator to the existence of abnormal activity in a video.  In a scene with many different activities occurring, it may not be immediately obvious to a human operator what the anomalous activity is if there is no spatial localization of the anomaly.  Indicating the anomaly with a bounding box, or highlighting an anomalous region of pixels allows the human operator to understand what anomaly has been detected.  Furthermore, another useful function of a VAD system is to provide a human-understandable explanation for a detected anomaly.  An explanation generally requires the system to know what regions or objects in a frame are detected as anomalous which requires spatially localizing anomalies.

Partially because many of the single-scene VAD datasets do not contain instances of the more difficult context-dependent and location-dependent anomaly types, single-scene VAD is viewed as simpler than multi-scene VAD and has seen a sharp decline in research papers that focus on it.  Looking at papers published in some of the major computer vision conferences over the last two years shows clear trends in the focus of research in video anomaly detection.  As shown in Figure \ref{fig:survey} (see also Table \ref{tab:survey} in the Appendix), there were 46 papers published in CVPR, ICCV, ECCV, NeurIPS and WACV in 2024 and 2025 that were focused specifically on video anomaly detection \cite{mulde2024,clap2024,ovvad2024,prompt2024,harnessing2024,guidance2024,causation2024,maskedAE2024,multiscale2024,pi2025,coding2025,track2025,anomize2025,vera2025,holmes2025,rules2024,crossdomain2024,interleave2024,fedvad2024,normality2024,mixture2025,keypoint2025,tokens2025,autoreg2025,realtime2024,c2fpl2024,oectst2024,holistic2024,jigsaw2025,egocentric2025,missiongnn2025,distill2025,pvvtt2025,score2025,review2024,multidomain2024,hawk2024,panda2025,holistic2025,event2025,a2seek2025,vadr12025,cooccurrence2025,frameshield2025,monitor2025,vadtree2025}.  Of those papers, all of them described methods intended mainly for multi-scene scenarios although a few also tested on single-scene datasets.  About 43\% of the papers (20 out of 46) used the weakly-supervised, multi-scene version of VAD.  None of the papers in those conferences described algorithms that built location-dependent models that are needed for single-scene video anomaly detection.  Furthermore, only 5 out of the 46 papers evaluated their VAD algorithm on spatial localization accuracy.  All of the rest used the frame-level criterion which only evaluates temporal localization of anomalies. Furthermore, 16 out of 46 papers used LLMs for their built-in notion of normality instead of learning normality solely from the training set.

\section{Limitations of Current VAD Paradigms}

The recent shift toward general models and LLM-driven approaches has transformed the way the research community conceptualizes video anomaly detection. While these methods promise scalability, they also introduce methodological and interpretive limitations that affect both scientific rigor and practical reliability. These limitations can be analyzed through the following observations.

\paragraph{(1) Methods using general models intended to apply across different scenes reduce the connection between anomaly detection and scene-specific normality.} 
These formulations assume normal activity is the same in every scene. While they perform well when activities like burglary or falling down are consistently treated as anomalies, this assumption does not hold in general. General models also do not readily capture location-dependent anomalies since locations are specific to a single scene. Loitering, for example, might be normal behavior in some locations, but suspicious in others. General models therefore abstract away the scene-specific context that defines anomalous behavior.

\paragraph{(2) Foundation-model priors and current weakly-supervised VAD practices introduce semantic biases that shift VAD toward action recognition rather than anomaly detection.}
Large pretrained models often contain extensive examples of common anomalous events such as fire, burglary, fighting, or accidents. As a result, VAD systems built on these priors tend to respond to semantically salient actions that match pretraining categories rather than to deviations from normal behavior in the target scene. This behavior blurs the distinction between anomaly detection and category-level action recognition. Pretrained semantic priors therefore bias VAD systems toward action-category responses rather than context-conditioned anomaly reasoning. Furthermore, weakly-supervised models trained on benchmark datasets are typically not evaluated on unseen anomaly types, since standard train-test splits contain the same set of anomaly types. This limits the ability to assess bias toward seen anomaly classes that may be learned during training.

\paragraph{(3) The emphasis on multi-scene generalization discourages spatial localization and reduces interpretability.}
Most weakly supervised pipelines produce only frame-level anomaly scores, discarding spatial cues needed for verifying detections or supporting practical decision-making. Without localization, systems cannot provide meaningful explanations or distinguish genuine anomalies from dataset artifacts. The lack of spatial grounding therefore limits interpretability and practical utility.

\paragraph{(4) The dependence on opaque pretrained foundation models limits transparency and reproducibility.}
Modern VAD pipelines routinely adopt large pretrained models whose training data are undisclosed. Because the content and scope of these datasets are unknown, it is not possible to determine whether evaluation data were seen or approximated during pretraining. This lack of transparency complicates fair evaluation and interpretability. Opaque pretraining data therefore limit reliable assessment of model generalization.

Together, these issues suggest that the current trajectory, while technologically innovative, may weaken the conceptual foundations of video anomaly detection. Addressing these limitations requires increased attention to contextual understanding, transparent evaluation, and scene-specific reasoning, rather than relying primarily on multi-scene generalization and pretrained semantic priors. In the next section, we demonstrate these effects through targeted empirical and visual analyses designed to highlight their practical impact.

\section{Experiments}

\definecolor{lblue}{RGB}{240,240,255}     
\definecolor{lgreen}{RGB}{245,255,245}  
\definecolor{lyellow}{RGB}{255,250,230}  
\definecolor{lred}{RGB}{255,230,230}      

\begin{table}
  \centering
  \begin{tabular}{ c | c  c  c  c }
    Method & UCF-Crime & StreetScene &  RPD   \\
    \hline
    \rowcolor{lblue}
    VADTree & 0.85 & 0.49 & 0.36   \\
    \rowcolor{lblue}
    EventVAD & 0.82 & 0.53 & 0.29   \\
    \rowcolor{lblue}
    LAVAD & 0.80 & 0.56 & 0.24   \\
    \rowcolor{lyellow}
    VERA & 0.86 & 0.00 & 0.86  \\
    \rowcolor{lgreen}
    Contextual GMM & n/a & 0.67 & n/a   \\
    \rowcolor{lgreen}
    MLLM-EVAD  & n/a & 0.67 & n/a   \\
    
  \end{tabular}
\caption{Comparison of video anomaly detection methods across multi-scene and single-scene benchmarks using the frame-level AUC as the evaluation metric. Methods are grouped by supervision type: 
\colorbox{lblue}{\textcolor{black}{Training-free}}, 
\colorbox{lyellow}{\textcolor{black}{Weakly-supervised  multi-scene}}, and 
\colorbox{lgreen}{\textcolor{black}{Semi-supervised  single-scene}}. 
RPD denotes Relative Performance Drop, highlighting the deficit when transitioning from multi-scene to single-scene evaluation.}

  \label{tab:results}
  
\end{table}

\paragraph{Experimental Setup.}
To complement our conceptual analysis, we present a targeted empirical evaluation designed to assess how different classes of video anomaly detection methods behave when applied to challenging single-scene settings. Our goal is to examine the robustness of prevailing approaches when evaluated outside the regime they are primarily designed for.

We consider representative methods spanning three categories: training-free approaches, weakly-supervised multi-scene methods, and semi-supervised single-scene models. Performance is reported on UCF-Crime~\cite{sultani2018real}, a large multi-scene benchmark, and on StreetScene~\cite{ramachandra2020street}, a single-scene dataset that requires precise modeling of scene-specific normality. Results are reported using the common frame-level area under the curve (AUC) evaluation metric \cite{li2013anomaly}. To isolate robustness across settings, we additionally report Relative Performance Drop (RPD), which measures the degradation in performance when methods transition from multi-scene to single-scene evaluation. Results for UCF-Crime are taken from the respective publications for each method. For StreetScene, we implement the weakly-supervised and training-free methods and report results from the respective publications for the semi-supervised single-scene methods.

\paragraph{Results.}
The results in Table~\ref{tab:results} reveal a clear and consistent pattern across supervision paradigms. Training-free methods show a significant drop in accuracy when transitioning from multi-scene to single-scene evaluation. For example, VADTree~\cite{vadtree2025} achieves an AUC of 0.85 on UCF-Crime and 0.49 on StreetScene, corresponding to a relative performance drop (RPD) of 0.36, while EventVAD~\cite{eventvad2025} and LAVAD~\cite{harnessing2024} exhibit comparable degradation trends with RPD values of 0.29 and 0.24, respectively. These results suggest that training-free methods are more suited to detecting general classes of anomalies present in UCF-Crime (such as fighting, burglary, arson, etc.) than scene-specific anomalies found in StreetScene (such as jaywalking and illegal parking). We note that for the frame-level criterion, which measures the false positive rate versus correct detection rate of frames, an AUC of 0.5 corresponds to chance-level performance. Thus, training-free methods operate near chance-level accuracy in this setting (slightly above chance in the case of LAVAD). In comparison, semi-supervised single-scene methods achieve substantially stronger performance on the single-scene benchmark, with Contextual GMM~\cite{contextual2025} and MLLM-EVAD~\cite{mllm-evad2025} both reaching an AUC of 0.67 on StreetScene, consistent with their explicit modeling of scene-specific normal behavior.

In contrast, the weakly supervised multi-scene method VERA~\cite{vera2025} exhibits a pronounced degradation under single-scene evaluation. As shown in Table~\ref{tab:results}, VERA attains competitive performance on UCF-Crime with a frame-level AUC of 0.86 but degrades sharply on StreetScene, not detecting anomalies until a false positive rate of 1.0 is reached, resulting in an AUC of 0.0. This behavior arises because VERA outputs a binary decision for each frame (anomalous or normal) and classifies all frames in StreetScene as normal. Notably, this result is obtained under a favorable training configuration: during weakly supervised training, VERA is exposed to all anomalous videos present in the StreetScene test set, eliminating uncertainty about anomaly category coverage. Despite this advantage, VERA does not identify anomalies in the single-scene setting, indicating that its limitations stem not from missing semantic knowledge but from an inability to localize and reason about anomalies in a spatially and contextually grounded manner.

\paragraph{Analysis.}
In addition to their improved quantitative performance, semi-supervised single-scene methods differ fundamentally from multi-scene weakly-supervised and training-free approaches in their ability to provide spatially localized anomaly detections. As shown in Table~\ref{tab:results}, methods such as Contextual GMM and MLLM-EVAD not only outperform the weakly-supervised baseline on StreetScene by more than 10 percentage points in AUC, but also produce localization-aware outputs that identify where anomalous behavior occurs within the scene. This capability is critical in single-scene settings, where anomalies are often defined by their spatial context rather than by global appearance or action category. In contrast, both training-free and weakly-supervised methods operate at the frame level and lack mechanisms for spatial grounding, preventing them from distinguishing between contextually normal and abnormal instances of the same activity. The superior performance of single-scene methods on StreetScene therefore reflects not only better numerical scores, but also a closer alignment with the requirements of practical anomaly detection, where localization and interpretability are essential.

These findings empirically support the limitations discussed in Section~3: methods designed around weak supervision and multi-scene generalization struggle to capture the contextual and spatial constraints that define anomalies in practical deployments. Even when weakly-supervised models are granted explicit access to the anomaly types present at test time, they do not generalize effectively to settings where abnormality is defined by local structure rather than semantic category. While such approaches scale across heterogeneous datasets, they do not provide a reliable foundation for single-scene anomaly detection, where understanding local normality is essential. We emphasize that these results are intended to illustrate structural limitations rather than absolute performance rankings.

\section{Limitations and Future Directions}

While our analysis highlights several limitations of current VAD paradigms, it is important to recognize the scope and trade-offs of different approaches. In this section, we discuss these considerations and outline promising directions for future research.

A prevailing alternative perspective is that models of normal activity should be applicable across many different scenes to improve practicality. General models offer the advantage that they do not require retraining for each new environment, and they can effectively detect common anomaly categories such as fighting, burglary, or falling. These properties make multi-scene approaches useful in many settings where detecting common anomaly types is sufficient.

However, a general model cannot capture all aspects of anomaly detection, since what is anomalous in one scene may not be anomalous in another. This observation motivates the need for approaches that better model scene-specific activity and can detect not only common anomaly types but any activity that differs significantly from normal activity seen in training videos.  Surveillance cameras that are setup to observe only one particular scene are a very important use case for VAD systems.  For such scenarios, our analysis suggests that effective VAD systems can benefit from learning models from video of a single scene that emphasize the following properties: \textbf{ (i) unbiased, (ii) location and context dependent, (iii) efficient, (iv) extendable, and (v) explainable.}

By {\it unbiased}, we mean that the model is learned only from normal video of a scene without relying on predefined notions of what is normal or anomalous. Such assumptions may arise from predefined anomaly categories (such as fighting or falling) or from pretrained models with built-in semantic priors. Determining normal activity in a particular environment generally requires observing that scene directly.

A model for single-scene VAD can also benefit from being {\it location dependent} and {\it context dependent}. Location and context dependence mean that an activity cannot be judged to be normal or anomalous without considering where it occurs and under what conditions. Activities may be normal in some locations or contexts within a scene but anomalous in others. For example, whether a parked car is normal depends on its location, and whether a moving vehicle is normal depends on its context, such as whether it is on a roadway or in a restricted area.

Learning a model of normal activity from video should also be computationally {\it efficient}, meaning that it is practical to train on standard hardware within a reasonable time. Training deep network models on millions of frames for each new scene may be impractical, whereas approaches that leverage pretrained feature extractors (e.g., object detectors or trackers) can offer more efficient alternatives. Related to efficiency, models can benefit from being {\it extendable}, meaning that they can be updated as new normal data becomes available without requiring complete retraining.

For human operators to effectively use VAD systems, models should be {\it explainable}, providing interpretable insights into why an event is considered normal or anomalous. This requirement is not limited to multi-scene settings, but also applies to recent challenging single-scene datasets such as StreetScene \cite{ramachandra2020street} and ComplexVAD \cite{mumcu2025complexvad}, where understanding the spatial and contextual basis of anomalies is essential.

The efficiency challenge remains particularly important for single-scene methods, as they require training for each new environment. A key direction for future work is developing methods that can efficiently learn and update models of normal activity. Approaches that first extract higher-level representations, such as objects or motion patterns, may provide a promising alternative to training directly on raw video data.

Overall, these directions highlight opportunities for developing VAD systems that are better aligned with real-world deployment settings, where scene-specific reasoning, interpretability, and adaptability are essential.

\section{Conclusions}

The recent, almost exclusive, research focus on multi-scene methods for video anomaly detection has diverted attention from important and interesting problems that only occur in single-scene scenarios. This work demonstrates that general models which apply to multiple scenes are often not suitable for the scene-specific, location-specific and context-dependent anomaly types that are common in single-scene VAD scenarios.
A return of focus to single-scene models that are unbiased, location and context dependent, efficient, extendable and explainable will require researchers to solve important problems that have broad applicability in many practical surveillance scenarios.

\setlength{\bibsep}{4pt}
\bibliography{references}
\bibliographystyle{plainnat}

\newpage
\appendix
\onecolumn
\section{More details from survey of recent VAD papers}

Table \ref{tab:survey} gives the raw data that was visualized in Figure \ref{fig:survey}.
It shows an exclusive focus on multi-scene VAD methods which especially concentrate on weakly-supervised and training-free approaches.  None of the recent methods build a model that can detect location-dependent anomalies, and very few include an evaluation of spatial localization accuracy.

\begin{table*}[t]
\centering
\caption{Statistics of recent papers on video anomaly detection.}
\vspace{-.5em}
\label{tab:survey}
\resizebox{\linewidth}{!}{
\begin{tabular}{l c c c c c}
\toprule
 &   & \textbf{Num using weakly-supervised} & \textbf{Num using location-} & \textbf{Num using LLM's built-in} & \textbf{No. evaluating spatial}\\
\textbf{Conference} & \textbf{Num papers} & \textbf{multi-scene protocol} & \textbf{specific model} & \textbf{notion of normality} & \textbf{localization}\\
\midrule
CVPR 2025 & 6 & 4 & 0 & 2 & 1\\
ICCV 2025 & 4 & 2 & 0 & 1 & 1\\
WACV 2025 & 6 & 4 & 0 & 1 & 0\\
NeurIPS 2025 & 9 & 2 & 0 & 7 & 3\\
CVPR 2024 & 9 & 3 & 0 & 3 & 0\\
ECCV 2024 & 5 & 3 & 0 & 1 & 0\\
WACV 2024 & 4 & 2 & 0 & 0 & 0\\
NeurIPS 2024 & 3 & 0 & 0 & 1 & 0\\
\midrule
Total & 46 & 20 & 0 & 16 & 5\\
\bottomrule
\end{tabular}
}
\end{table*}

\end{document}